\documentclass{article}
\usepackage{PRIMEarxiv}

\usepackage[utf8]{inputenc} 
\usepackage[T1]{fontenc}    
\usepackage{hyperref}       
\usepackage{url}            
\usepackage{booktabs}       
\usepackage{amsfonts}       
\usepackage{nicefrac}       
\usepackage{microtype}      
\usepackage{fancyhdr}       
\usepackage{graphicx}       
\usepackage{amsmath}
\usepackage{amssymb}
\usepackage{cite}
\usepackage{booktabs}
\usepackage{subfigure}
\usepackage{multirow}
\usepackage[normalem]{ulem}
\usepackage{balance}
\usepackage{pifont}
\usepackage{adjustbox}
\usepackage{hyperref}       
\hypersetup{
        colorlinks = true,
        allcolors = magenta,
        breaklinks = true,
        linktocpage=true,
}
\graphicspath{{./Figures/}}
\def\x{{\mathbf x}}														
\def\y{{\mathbf y}}														
\def\W{{\mathbf W}}														
\def\ii{{\hat{\imath}}}												
\def\ij{{\hat{\jmath}}}												
\def\ik{{\hat{\kappa}}}												
\def\bC{\mathbb{C}}														
\def\bH{\mathbb{H}}														
\def\bO{\mathbb{O}}														
\def\bR{\mathbb{R}}														
\def\bS{\mathbb{S}}														
\def\bT{\mathbb{T}}														
\title{Demystifying the Hypercomplex: \\Inductive Biases in Hypercomplex Deep Learning
}

\author{
  Danilo~Comminiello$^\circ$\thanks{Corresponding author's email: danilo.comminiello@uniroma1.it. \\
	\textit{\underline{Citation}}: 
\textbf{Danilo Comminiello, Eleonora Grassucci, Danilo P. Mandic, Aurelio Uncini, ``Demystifying the Hypercomplex: Inductive Biases in Hypercomplex Deep Learning,'' \textit{IEEE Signal Processing Magazine}, 2024.}
}, Eleonora~Grassucci$^\circ$,  Danilo~P.~Mandic$^\dagger$, Aurelio~Uncini$^\circ$\\
  $^\circ$Dept. Information Engineering, Electronics, and Telecommunications, Sapienza University of Rome, Italy \\
  $^\dagger$Dept. Electrical and Electronic Engineering, Imperial College London, UK\\
}
\begin{document}
\maketitle
%
%
%
\begin{abstract}
Hypercomplex algebras have recently been gaining prominence in the field of deep learning owing to the advantages of their division algebras over real vector spaces and their superior results when dealing with multidimensional signals in real-world 3D and 4D paradigms.
This paper provides a foundational framework that serves as a roadmap for understanding why hypercomplex deep learning methods are so successful and how their potential can be exploited. Such a theoretical framework is described in terms of inductive bias, i.e., a collection of assumptions, properties, and constraints that are built into training algorithms to guide their learning process toward more efficient and accurate solutions. We show that it is possible to derive specific inductive biases in the hypercomplex domains, which extend complex numbers to encompass diverse numbers and data structures. These biases prove effective in managing the distinctive properties of these domains, as well as the complex structures of multidimensional and multimodal signals. This novel perspective for hypercomplex deep learning promises to both demystify this class of methods and clarify their potential, under a unifying framework, and in this way promotes hypercomplex models as viable alternatives to traditional real-valued deep learning for multidimensional signal processing.
\end{abstract}


%
%
%
%
%
\section{Introduction}
\label{sec:intro}
\subsection{Advantages of Hypercomplex Deep Learning}
Recent technological developments in big data analysis have increased the availability of multidimensional signals (e.g., 3D data, point clouds, image volumes), multichannel signals (e.g., sensor arrays, smart grids), and multimodal signals (e.g., medical imaging, remote sensing, audio-visual content). 
Traditional real-valued deep-learning analysis methods for such data allow for the extraction of significant information and the accomplishment of even very complex tasks. However, they also require the development of very large architectures, sometimes even hierarchical, with multiple branches and also involving information fusion strategies \cite{TookTNNLS2015, WangNEURIPS2020, NagraniNEURIPS2021}. 
%
The level of complexity is even larger when multidimensional signals come into play, as they often exhibit particular structures that arise from the inherent characteristics of the data and the physical processes that generate them \cite{TookTNNLS2015, SonthaliaNEURIPS2021, LiSPM2023}. 

Indeed, the most significant deep learning models are tailored for processing data in the Euclidean domains and then extended to tasks in higher-dimensional (real) vector spaces. However, such networks typically do not fully take into account the dimensionality and the inherent coupling of variables of multidimensional input signals. Consequently, such learning algorithms are often not closed under mathematical operations, such as rotation and translation, thus leading to poor generalization ability and poor results in complex tasks. 
A possible solution is provided by redefining deep learning models directly in the hypercomplex domains. In this way, data can be processed in the very domain they reside, thus helping to mitigate the brute force approaches of (real-valued) deep learning models which process multidimensional data as a set of unrelated components. Indeed, traditional deep learning models predominantly reside in the fields of real, $\mathbb{R}$, and complex, $\mathbb{C}$, numbers, which are not guaranteed to capture all the intricacies of certain signal types or complex tasks. Hypercomplex deep learning instead extends the capabilities of deep learning theory by incorporating hypercomplex numbers into neural network computations. This allows hypercomplex models to take advantage of:
\begin{itemize}
    \item  The multidimensional and highly correlated nature of input signals;
    \item The operations defined in the hypercomplex domain, which are derived from a set of algebraic rules, such as the Hamilton product.
\end{itemize}

\noindent The ability of hypercomplex deep learning to capture spatial relationships, complex rotations, and intricate patterns makes it a valuable approach for tasks that require a comprehensive physical understanding of data across multiple dimensions and orientations (i.e., domain knowledge).
%

\subsection{Need for Inductive Biases in Hypercomplex Domains -- Motivation and Goals of the Paper}
Traditional deep learning has been widely investigated in the recent literature to provide a clear and exhaustive explanation of why some models may bring impressive results in specific problems. In particular, several studies focused on the set of properties and theoretical assumptions, also known as inductive biases, that lead a deep learning model to guide the learning process toward effective solutions \cite{Mitchell1980, Goyal2022, GoldblumARXIV2023}. However, similar analyses regarding the study of the fundamental assumptions that make hypercomplex models suitable for multidimensional signals are lacking in the literature. 

In this paper, we provide a novel comprehensive analysis of inductive biases for hypercomplex deep learning in order to demystify the learning algorithms defined in the hypercomplex domains, while also explaining the assumptions that can be exploited in specific multidimensional signal processing applications.
%
%
In particular, we describe potential inductive biases that mainly rely on the dimensionality of the input data and on domain-specific algebraic operations as well as geometric transformations. This set of constraints and assumptions constitutes the inductive bias for this type of hypercomplex model that leads to four crucial advantages over the traditional real-valued deep learning models:
\begin{itemize}
    \item Outstanding saving on computational costs, due to a reduction of the parameters up to $1/n$ (where $n$ can be treated as the dimensionality of the input signals);
    \item Greater ability to generalize models concerning unknown data even in very complex problems;
    \item Compact representation of real-valued multidimensional signals in single hypercomplex-valued entities;
    \item Ability to model sets of transformations in higher-dimensional spaces.
\end{itemize}

\noindent We show that these biases impose constraints on the relationships and interactions among the individual entities in the learning process. This equips a hypercomplex deep learning model with the ability to better generalize on the unknown data compared to its real-valued counterpart, while also saving a significant amount of model parameters.


\subsection{Paper Organization}

The rest of the paper is organized as follows. In Section~\ref{sec:hypdom}, we provide an analytical perspective on how to incorporate algebraic properties into learning operations. Section~\ref{sec:hyperinductive} examines the set of inductive biases that hypercomplex model can benefit from. Section~\ref{sec:phnninductive} explains how parameterized hypercomplex models force some inductive bias advantages beyond the strict algebraic rules while generalizing their benefits to a wide range of multidimensional signals. Finally, in Section~\ref{sec:applications}, we show how to leverage the hypercomplex inductive biases in several applications. Conclusion and future directions are drawn in Section~\ref{sec:conclusion}.



%
%
%
%
%
\section{Learning in Hypercomplex Domains}
\label{sec:hypdom}
\subsection{Hypercompex Algebras for Multidimensional Signals} 
\label{subs:hypalg}
Hypercomplex deep learning models can be defined according to a hypercomplex number system, characterized by a set of algebra rules that define mathematical operations. Hypercomplex algebras belong to the family of Cayley-Dickson algebras and can be distinguished based on the dimensionality of order 
\begin{equation}
    n=2^m, \qquad \text{with } m \in \mathbb N. 
    \label{eq:CDconstraint}
\end{equation}

\begin{figure}[t!]
    \centering
    \includegraphics[width=\textwidth]{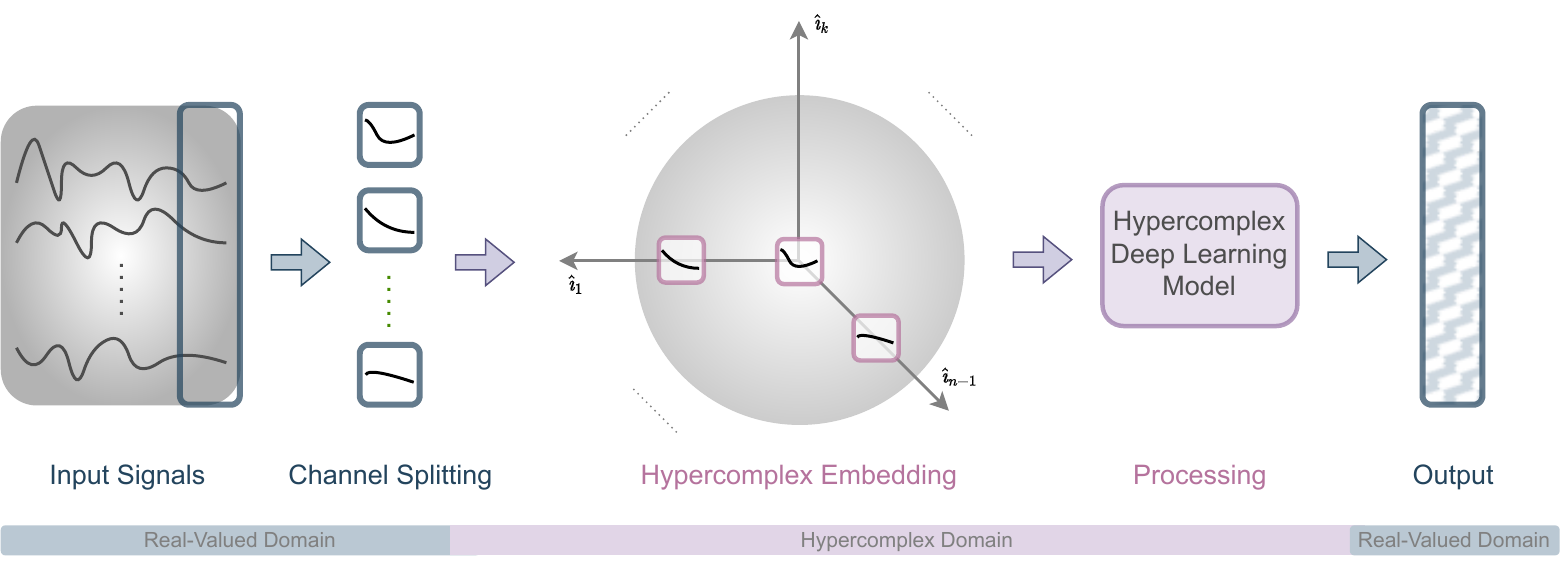}
    \caption{Workflow of hypercomplex deep learning. Individual channels of any multidimensional signal are split in the real-valued domain and then embedded in the hypercomplex domain where the hypercomplex-valued input is processed by a hypercomplex deep learning model. The latter will result in a real-valued input, whose nature (e.g., signals, label) will depend on the specific task (e.g., regression, classification, segmentation, synthesis, or other).}
    \label{fig:hypelearn}
\end{figure}

\noindent Indeed, we can derive the complex-valued domain, $\bC$, for $n=2$; quaternion domain, $\bH$, and tessarine domain, $\bT$, for $n=4$; octonions, $\bO$, for $n=8$; sedenions, $\bS$, for $n=16$; and so on. This wealth of hypercomplex domains makes them suitable for handling and processing multidimensional signals, whose dimensional order satisfies the Cayley-Dickson constraint of \eqref{eq:CDconstraint}. Each dimensional component of a signal can in fact be represented on an axis of a hypercomplex domain, as also shown in Fig.~\ref{fig:hypelearn}.

A generic multidimensional input signal of dimensionality $n$ can be represented in the hypercomplex domain as
\begin{equation}
    x = x_0 + x_1 \ii_1 + \ldots + x_{n-1} \ii_{n-1}, 
\label{eq:hinput}
\end{equation}

\noindent where $x$ is a hypercomplex-valued signal, $x_0, \ldots, x_{n-1}$ are the real-valued components (representing each dimension of the input), and $\ii_1, \ldots, \ii_{n-1}$ are the corresponding imaginary units. Observe that a real number can be expressed through \eqref{eq:hinput} by setting $n=1$, as $x = x_0 + 0\ii_i+ \ldots + 0\ii_n$. A hypercomplex number with zero real part $x_0$ and only the imaginary components, $x = 0 + x_1 \ii_1 + \ldots + x_n \ii_{n-1}$, is termed \textit{pure}. Hypercomplex numbers are endowed with a component-wise addition
\begin{equation}
    x + y = (x_0+y_0) + (x_1+y_1) \ii_i + \ldots + (x_{n-1}+y_{n-1}) \ii_n,
\end{equation}

\noindent and a product for any real-valued scalar $\alpha$
\begin{equation}
    \alpha h = \alpha x_0 + \alpha x_1 \ii_1 + \ldots + \alpha x_{n-1} \ii_{n-1}.
\end{equation}

On the contrary, the vector multiplication of hypercomplex numbers obeys completely different rules with respect to summation and subtraction. Indeed, it has to take into account the interactions among the imaginary units and their non-commutativity under this operation. 

All the imaginary components of a hypercomplex algebra in the Cayley-Dickson formulation anti-commute and satisfy the property whereby $\ii_l^2 = -1$, for $l = 1, \ldots, n-1$. When $m \geq 3$ in \eqref{eq:CDconstraint}, we have algebras with dimensionality larger than $n = 8$ that are non-associative. Also, all the resulting algebras for $m \geq 4$, and consequently $n \geq 16$, have zero-divisors. Each time $m$ increases, we lose an algebraic symmetry (e.g., commutativity for quaternions, associativity for octonions, norm-multiplicativity for sedonions), as Table~\ref{tab:algebras} shows.

One of the most popular examples of hypercomplex algebra is quaternion algebra, which is employed in this work to demonstrate practical applications. 
%
A quaternion variable is defined as $x \in \bH = x_0 + x_1 \ii + x_2 \ij + x_3 \ik$, where $x_i, \; i \in \{0,1,2,3\}$ are real-valued coefficients and $\ii, \ij, \ik$ are the imaginary units, whereby $\ii^2 = \ij^2 = \ik^2 = -1$ and $\ii \ij = - \ij \ii ; \; \ij \ik = - \ik \ij ; \; \ik \ii = - \ii \ik$. Consequently, one of the most peculiar and important operations in the quaternion domain is the quaternion product, also known as the Hamilton product, which defines the product of two quaternions by taking into account the associativity and the non-commutativity properties of quaternions. It is worth noting that multiplications are differently shaped as the domain changes. For example, 
quaternions and tessarines may have different multiplication rules, as shown in Tab.~\ref{tab:algebras}.
%

%
%
%
%
\begin{table*}[t!]
{\tiny{
\centering
\caption{Algebraic convolution and properties for domains.}
\label{tab:algebras}
\begin{tabular}{@{}llllccccc@{}}
\toprule
Domain          & Input & Weight & Multiplication($\otimes$) $\backslash$ Convolution($*$) & \multicolumn{4}{c}{Multiplication Properties} \\ \midrule
& & & & \rotatebox[]{90}{Comm.} & \rotatebox[]{90}{Ass.} & \rotatebox[]{90}{Alt.} & \rotatebox[]{90}{Power ass.} \\ \midrule
Real            & $\x$ & $\W$ & $\W$ $\times\backslash *$ $\x$ & \ding{51} & \ding{51} & \ding{51} & \ding{51} \\
\midrule[.1em]
\multirow{2}{*}{Complex}         & \!$\begin{aligned}[t]&\x_0 \\+&\x_1\ii_1 \end{aligned}$& \!$\begin{aligned}[t]&\W_0 \\+&\W_1\ii_1 \end{aligned}$ & \multirow{2}{*}{$\left[\begin{array}{*{10}c} \W_0 & -\W_1 \\
                                                    \W_1 & \W_0 \end{array} \right] \otimes \backslash * \left[\begin{array}{*{10}c}
                                                         \x_0\\
                                                         \x_1 \end{array}  \right]$} & \multirow{2}{*}{\ding{51}} & \multirow{2}{*}{\ding{51}} & \multirow{2}{*}{\ding{51}} & \multirow{2}{*}{\ding{51}} \\
\midrule[.1em]
\multirow{5}{*}{\rotatebox{90}{Quaternion}}      & \!$\begin{aligned}[t]&\x_0 \\+&\x_1\ii_1 \\+&\x_2\ii_2 \\+&\x_3\ii_3 \end{aligned}$ & \!$\begin{aligned}[t]&\W_0 \\+&\W_1\ii_1 \\+&\W_2\ii_2 \\+&\W_3\ii_3 \end{aligned}$ & \multirow{5}{*}{$\left[\begin{array}{*{10}c} \W_0 & -\W_1 & -\W_2 & -\W_3 \\
        \W_1 & \W_0 & -\W_3 & \W_2 \\
        \W_2 & \W_3 & \W_0 & -\W_1\\
        \W_3 & -\W_2 & \W_1 & \W_0 \end{array} \right] \otimes \backslash * \left[\begin{array}{*{10}c}
                                                         \x_0\\
                                                         \x_1 \\
                                                         \x_2 \\
                                                         \x_3 \end{array}  \right] $} & \multirow{5}{*}{\ding{55}} & \multirow{5}{*}{\ding{51}} & \multirow{5}{*}{\ding{51}} & \multirow{5}{*}{\ding{51}} \\
\midrule[.1em]
\multirow{5}{*}{\rotatebox{90}{Tessarine}} &  \!$\begin{aligned}[t]&\x_0 \\+&\x_1\ii_1 \\+&\x_2\ii_2 \\+&\x_3\ii_3 \end{aligned}$ & \!$\begin{aligned}[t]&\W_0 \\+&\W_1\ii_1 \\+&\W_2\ii_2 \\+&\W_3\ii_3 \end{aligned}$ & \multirow{5}{*}{$\left[\begin{array}{*{10}c} \W_0 & -\W_1 & \W_2 & -\W_3 \\
        \W_1 & \W_0 & \W_3 & \W_2 \\
        \W_2 & -\W_3 & \W_0 & -\W_1\\
        \W_3 & \W_2 & \W_1 & \W_0 \end{array} \right] \otimes \backslash * \left[\begin{array}{*{10}c}
                                                         \x_0\\
                                                         \x_1 \\
                                                         \x_2 \\
                                                         \x_3 \end{array}  \right] $} & \multirow{5}{*}{\ding{51}} & \multirow{5}{*}{\ding{51}} & \multirow{5}{*}{\ding{51}} & \multirow{5}{*}{\ding{51}} \\
\midrule[.1em]
\multirow{10}{*}{\rotatebox{90}{Dual Quaternion}} & \!$\begin{aligned}[t]&\x_0 \\+&\x_1\ii_1 \\+&\x_2\ii_2 10\\+&\x_3\ii_3\\ +&\epsilon(\x_0 \\+&\x_1\ii_1 \\+&\x_2\ii_2 \\+&\x_3\ii_3)\end{aligned}$ & \!$\begin{aligned}[t] &\W_0 \\+&\W_1\ii_1 \\+&\W_2\ii_2 \\+&\W_3\ii_3\\ +&\epsilon(\W_0 \\+&\W_1\ii_1 \\+&\W_2\ii_2 \\+&\W_3\ii_3)\end{aligned}$ & \multirow{10}{*}{$\left[\begin{array}{*{10}c} \W_0 & -\W_1 & -\W_2 & -\W_3 & \mathbf{0} & \mathbf{0} & \mathbf{0} & \mathbf{0} \\
        \W_1 & \W_0 & -\W_3 & \W_2 & \mathbf{0} & \mathbf{0} & \mathbf{0} & \mathbf{0} \\
        \W_2 & \W_3 & \W_0 & -\W_1 & \mathbf{0} & \mathbf{0} & \mathbf{0} & \mathbf{0} \\
        \W_3 & -\W_2 & \W_1 & \W_0 & \mathbf{0} & \mathbf{0} & \mathbf{0} & \mathbf{0} \\
        \W_4 & -\W_5 & -\W_6 & -\W_7 & \W_0 & -\W_1 & -\W_2 & -\W_3 \\
        \W_5 & \W_4 & -\W_7 & \W_6 & \W_1 & \W_0 & -\W_3 & \W_2 \\
        \W_6 & \W_7 & \W_4 & -\W_5 & \W_2 & \W_3 & \W_0 & -\W_1 \\
        \W_7 & -\W_6 & \W_5 & \W_4 & \W_3 & -\W_2 & \W_1 & \W_0  
        \end{array} \right] \otimes \backslash * \left[\begin{array}{*{10}c}
                                                         \x_0\\
                                                         \x_1 \\
                                                         \x_2 \\
                                                         \x_3 \\
                                                         \x_4 \\
                                                         \x_5 \\
                                                         \x_6 \\
                                                         \x_7 \end{array}  \right] $} & \multirow{10}{*}{\ding{55}} & \multirow{10}{*}{\ding{51}} & \multirow{10}{*}{\ding{51}} & \multirow{10}{*}{\ding{51}} \\
\midrule[.1em]
\multirow{10}{*}{\rotatebox{90}{Octonion}} & \!$\begin{aligned}[t]&\x_0 \\+&\x_1\ii_1 \\+&\x_2\ii_2 \\+&\x_3\ii_3\\ +&\x_4\ii_4 \\+&\x_5\ii_5 \\+&\x_6\ii_6 \\+&\x_7\ii_7\end{aligned}$ & \!$\begin{aligned}[t]&\W_0 \\+&\W_1\ii_1 \\+&\W_2\ii_2 \\+&\W_3\ii_3\\ +&\W_4\ii_4 \\+&\W_5\ii_5 \\+&\W_6\ii_6 \\+&\W_7\ii_7\end{aligned}$ & \multirow{10}{*}{$\left[\begin{array}{*{10}c} \W_0 & -\W_1 & -\W_2 & -\W_3 & -\W_4 & -\W_5 & -\W_6 & -\W_7 \\
        \W_1 & \W_0 & -\W_3 & \W_2 & -\W_5 & \W_4 & \W_7 & -\W_6 \\
        \W_2 & \W_3 & \W_0 & -\W_1 & -\W_6 & -\W_7 & \W_4 & \W_5 \\
        \W_3 & -\W_2 & \W_1 & \W_0 & -\W_7 & \W_6 & -\W_5 & \W_4 \\
        \W_4 & \W_5 & \W_6 & \W_7 & \W_0 & -\W_1 & -\W_2 & -\W_3 \\
        \W_5 & -\W_4 & \W_7 & -\W_6 & \W_1 & \W_0 & \W_3 & -\W_2 \\
        \W_6 & -\W_7 & -\W_4 & \W_5 & \W_2 & -\W_3 & \W_0 & \W_1 \\
        \W_7 & \W_6 & -\W_5 & -\W_4 & \W_3 & \W_2 & -\W_1 & \W_0  
        \end{array} \right] \otimes \backslash * \left[\begin{array}{*{10}c}
                                                         \x_0\\
                                                         \x_1 \\
                                                         \x_2 \\
                                                         \x_3 \\
                                                         \x_4 \\
                                                         \x_5 \\
                                                         \x_6 \\
                                                         \x_7 \end{array}  \right] $} & \multirow{10}{*}{\ding{55}} & \multirow{10}{*}{\ding{55}} & \multirow{10}{*}{\ding{51}} & \multirow{10}{*}{\ding{51}} \\ \bottomrule
\end{tabular}
}}\normalsize
\end{table*}

%
\subsection{Hypercomplex Deep Learning Fundamentals} 
\label{subs:hdl}
\begin{figure}[t]
    \centering
    \includegraphics[width=0.65\textwidth]{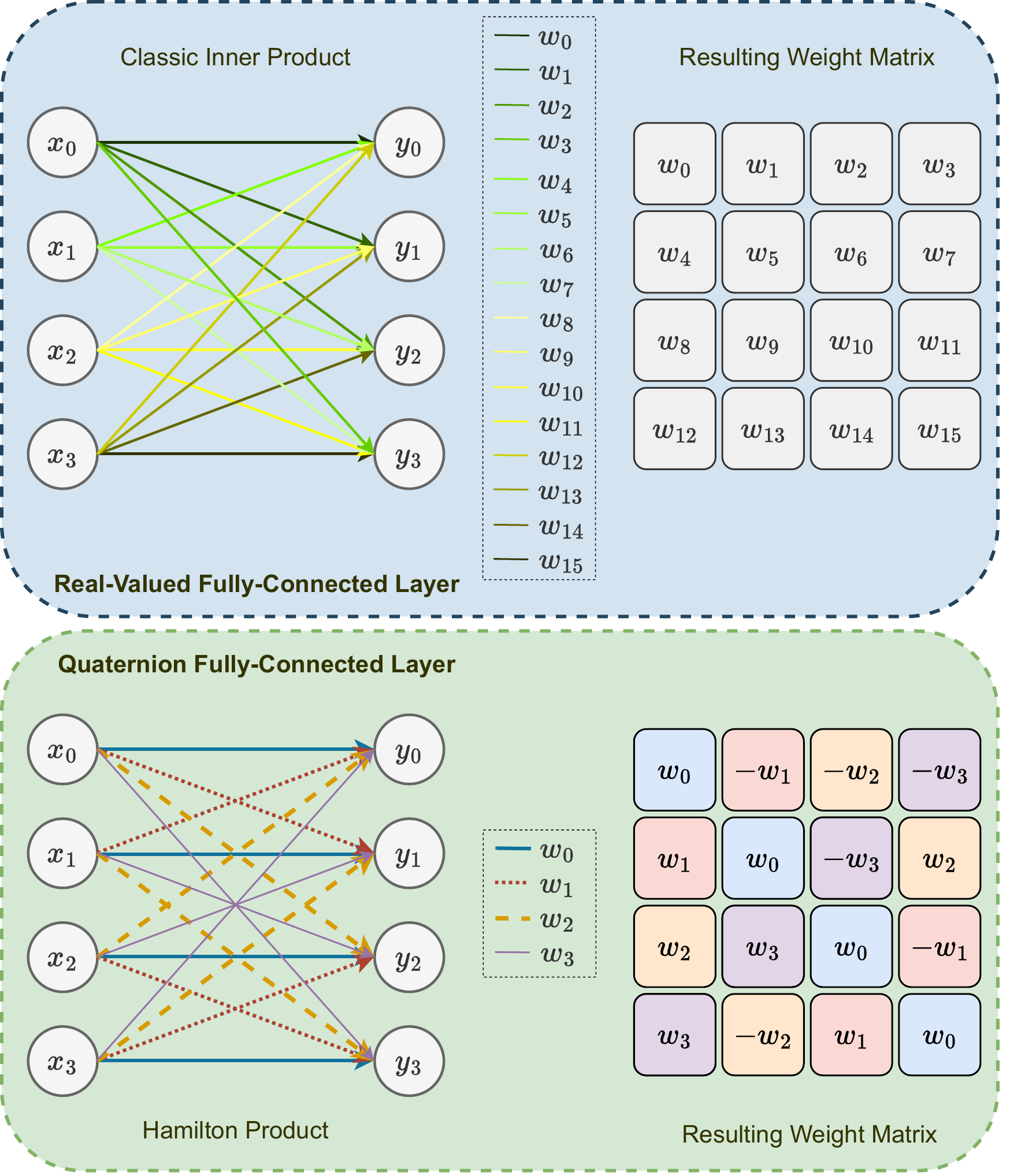}
    \caption{The Hamilton product is a clear example of relational inductive bias in the quaternion domain. Indeed, in the quaternion layer, only 4 weights are defined, which multiply each component of the input. The weight matrix is built by reusing these 4 weights while exhibiting the same size as a classic real-valued weight matrix. In this example, the Hamilton product involves $1/4$ of the parameters with respect to the real-valued networks.}
    \label{fig:qlayer}
\end{figure}

%
Hypercomplex deep learning overcomes the limitations of real-valued deep learning methods by exploiting the intrinsic algebraic properties that define the rules of learning. 
According to the specific hypercomplex algebraic system considered, it is possible to define different learning operations, which establish the relational rules and the interactions between the various imaginary units. Relational rules, especially related to the product between two hypercomplex numbers, such as vector multiplications, deeply affect the definition of the learning operations. 
Indeed, learning layers are usually defined as elements of the hypercomplex domain in which the neural network operates, and the operations of multiplication, convolution, and attention, among others, obey domain rules. Therefore, for any hypercomplex input $\x$, weight matrix $\W$ and bias $\mathbf{b}$, hypercomplex fully connected (HFC) layers can be defined as:
\begin{equation}
\label{eq:fc_layer}
    \mathbf{y} = \text{HFC}(\mathbf{x}) = \phi(\mathbf{W} \otimes \mathbf{x} + \mathbf{b}),
\end{equation}

\noindent and hypercomplex convolutional layers (HConv) as:
\begin{equation}
\label{eq:conv_layer}
    \mathbf{y} = \text{HConv}(\mathbf{x}) = \phi(\mathbf{W} * \mathbf{x} + \mathbf{b}),
\end{equation}

\noindent where $\phi$ is a nonlinear activation function, and the operations of multiplication $(\otimes)$ and convolution $(*)$ follow domain-specific rules and organize weights and filters as shown in Tab.~\ref{tab:algebras}
in order to take into account the commutativity and associativity properties. A graphical representation of the HFC layer in the case of the quaternion domain is shown in Fig.~\ref{fig:qlayer}.

Similarly, hypercomplex neural networks can be equipped with hypercomplex attention layers that, owing to the properties of the hypercomplex convolution (see Tab.~\ref{tab:algebras}),
jointly exploit local spatial information and long-range relationships \cite{Yang2023QTN}. Assuming an input $\x$, the hypercomplex self-attention layer follows:
\begin{equation}
\begin{aligned}
    \mathbf{h} &= \text{HConv}(\x), \\
    \text{HAtt} &= \text{HConv}_{1\times1}(\text{HConv}(\text{Concat}(\mathbf{h}, \mathbf{h}))), \\
    \mathbf{y} &= \text{HAtt} \cdot \x.
\end{aligned}
\end{equation}

Also, the aggregation operator of graph convolutional networks can be defined in hypercomplex domains considering the Hamilton product for quaternions \cite{Nguyen2021QGNN} and its extension to the other Cayley-Dickson algebra domains, as shown in Tab.~\ref{tab:algebras}.
Assuming a graph $\mathcal{G} = (\mathcal{V}, \mathcal{E})$, where $v \in \mathcal{V}$ is a node within the set of nodes $\mathcal{V}$ with the set of neighbors $\mathcal{N}_v$, and $v, u \in \mathcal{E}$ is an edge in the set of edges $\mathcal{E}$ between the node $v$ and the node $u$, the aggregation operator for the layer $l+1$ is defined as:
\begin{equation}
    \mathbf{h}_v^{l+1} = \phi \left( \Sigma_{u \in \mathcal{N}_v \bigcup\{v\}} a_{v,u} \W^l \otimes \mathbf{h}_u^l \right), \qquad \forall v \in \mathcal{V},
\end{equation}

\noindent whereby $a_{v,u}$ is a constant edge between the two nodes in the normalized adjacency matrix \cite{Nguyen2021QGNN}.

\subsection{Hypercomplex Correlation Learning}
\label{subs:hcorlearn}
\begin{figure}
    \centering
    \includegraphics[width=0.75\textwidth]{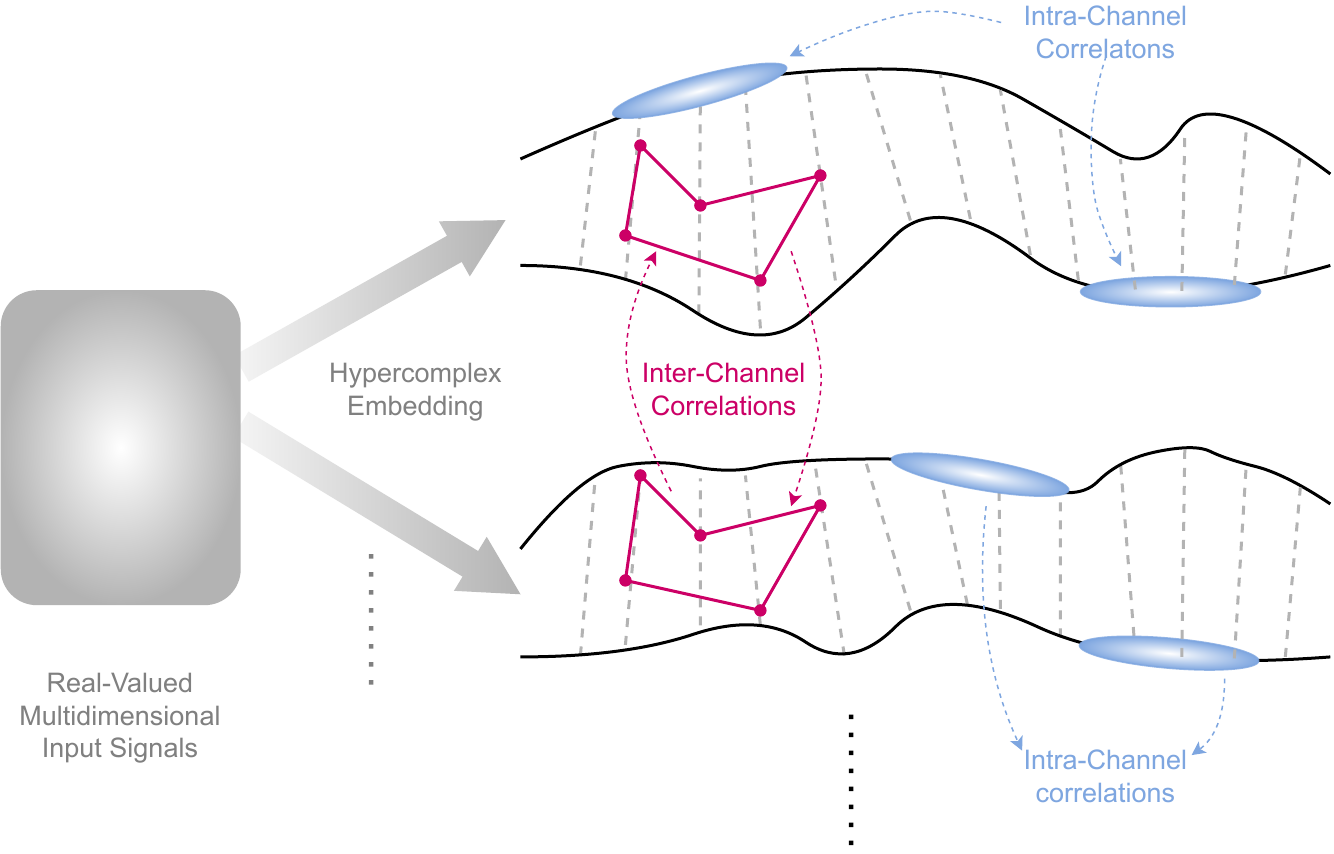}
    \caption{Each channel of the multidimensional input (i.e., $x_0, x_1, \ldots, x_{n-1}$) is embedded in hypercomplex axes where global and local features, related respectively to intra-channel and inter-channel correlations, are captured by learning algorithms. This favorable property is a major assumption in the definition of dimensionality bias.}
    \label{fig:hycor}
\end{figure}

\begin{figure}
    \centering
    \includegraphics[width=0.4\textwidth]{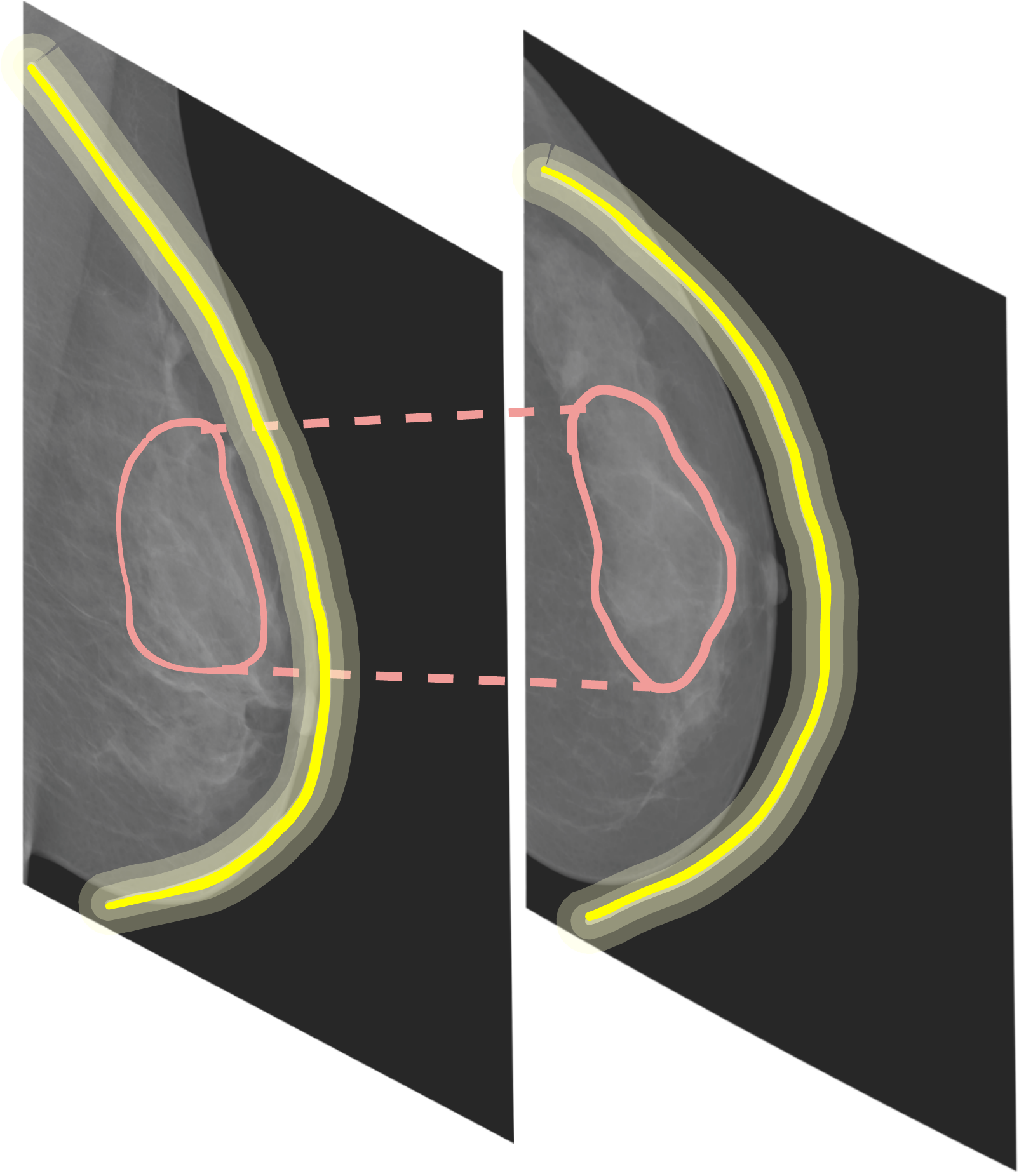}
    \caption{Intra-channel correlations are intra-view features such as the shape of the breast (in yellow), while inter-channel correlations are inter-view features such as textures (red). Original image in \cite{LopezARXIV2023}.}
    \label{fig:mamm_relations}
\end{figure}

The definition of hypercomplex neural network layers allows for better capturing inter-channel and intra-channel correlations compared to real-valued networks. Indeed, in hypercomplex neural networks, input channels can be thought of as a hypercomplex entity, whose components capture different aspects of the data related to inter- and intra-channel correlations, as also depicted in Fig.~\ref{fig:hycor} for a generic type of signals and more in practice in Fig.~\ref{fig:mamm_relations} for the case of multi-view mammographies.

The \textit{intra-channel correlation} pertains to the relationships between the same channel across different layers of the network. It involves the propagation of information within a specific channel over the network depth. This propagation can help refine and abstract features at different levels of representation. Intra-channel correlations help grasp global features, thus forward passes in hypercomplex layers act like low-pass filters highlighting shapes or curvatures in input signals. 
The \textit{inter-channel correlation} refers to the interactions and dependencies between different channels within the same layer, and this is peculiar in hypercomplex processing. For instance, in image processing, different channels could correspond to different color channels (e.g., red, green, blue) or feature maps learned from previous layers. Inter-channel interactions enable the network to capture complex feature relationships, enhance feature extraction by minimizing intra-ference \cite{TookTNNLS2015}, and provide a richer representation of the input data. In particular, these correlations capture local features so that a hypercomplex layer is likely to apply a high-pass filter. For image inputs for instance, this allows the network to emphasize color contrasts, edges, and textures effectively. 
Through intra-channel correlations, local features can evolve and interact across layers, thus resulting in increasingly abstract and discriminative representations. 

The simultaneous extraction of global and local features makes hypercomplex neural networks a powerful deep learning approach for several kinds of complex tasks. In fact, both inter-channel and intra-channel correlations contribute to the effectiveness of hypercomplex neural networks in capturing complex patterns, spatial relationships, and features invariant to geometric transformations, and thus favoring the definition of inductive biases in the hypercomplex domain.

\section{Inductive Biases in the Hypercomplex Domains} 
\label{sec:hyperinductive}
%
%
%
%
\subsection{Extending Biases from Real to Hypercomplex Domains} 
The inductive bias is a crucial concept in deep learning as it helps models to generalize well from the training data to unseen examples. Deep learning models make informed decisions based on the signals they process, in the same way humans rely on prior experience and knowledge to infer about new situations.

Inductive biases can be distinguished based on several characteristics. For instance, it is possible to categorize inductive biases on the basis of relational and non-relational assumptions \cite{BattagliaARXIV2018}. \textit{Relational inductive biases} focus on capturing and understanding relationships and interactions between different entities within the data. They are particularly relevant when dealing with structured or interconnected attributes, such as connectivity of neurons or neural blocks, topological architectures, social networks, knowledge bases, and also signals that exhibit a multimodal nature. Relational biases allow models to encode complex dependencies and interactions between entities. This is crucial in scenarios where the relationships between data points hold valuable information. Relational biases also facilitate higher-level cognition, reasoning and inference \cite{Goyal2022}. They can assist in transferring learned knowledge from one domain to another, as well as in continual learning and meta-learning scenarios \cite{Goyal2022, BahdanauICLR2019}. 
If a model grasps underlying relationships, it can adapt its knowledge to a new domain that shares similar structural characteristics. On the other hand, \textit{non-relational inductive biases} include a set of assumptions on signals and models that do not rely on their own relationships. These biases can be used to transform signal representations or to optimize and regularize loss functions, without affecting inherent relationships in signals and models. For instance, models like large language models (e.g., GPT) encode semantic meaning without explicitly representing relational connections. Examples of non-relational assumptions are represented by activation functions, regularization terms in the loss function, dropout, and batch normalization, among others.

Hypercomplex deep learning not only inherits all the inductive biases valid for the real-valued case, but extends them by exploiting some domain-specific algebraic and geometric definitions. Of particular significance above all are algebraic and geometric biases which fully complement the set of assumptions of real-valued deep learning models. Moreover, other biases defined in the real-valued domain, like the regularization bias, are extended with additional assumptions. We shall now introduce the most significant inductive biases that leverage the properties of hypercomplex algebras to improve the generalization ability of deep learning and develop state-of-the-art hypercomplex-valued deep learning models.
%

%
%
%
%
\subsection{Dimensionality Bias}
Relational assumptions may be imposed on the input signals to promote inductive bias. In the hypercomplex domains, this property can be straightforwardly exploited, and even stressed, due to the multidimensional nature of input signals. 
Indeed, neighboring samples in a multidimensional signal often exhibit strong correlations due to global features. This means that the values of nearby data points are likely to be similar or related in some way. For example, in an image, adjacent pixels often represent parts of the same object or scene, leading to \textit{spatial coherence}. Also, consecutive samples in time series data or videos may be correlated due to underlying \textit{temporal dependencies}. This is seen in stock market data, where consecutive stock prices are often influenced by past prices, or in video sequences, where consecutive frames exhibit smooth transitions. 
Besides temporal or spatial correlations, multidimensional signals, when analyzed in the frequency domain, may reveal local relations, such as patterns, textures, or structures that provide valuable insights into the contents \cite{EllTIP2007, ComminielloICASSP2019a}. 
This involves defining hierarchical structures, where smaller-scale patterns combine into larger-scale structures. In such situations, representing signals in a more compact form allows for separating different scales of details.

Such complex relations are all the more evident and structured the higher the dimensionality of the input signal. This assumption represents the \textit{dimensionality bias}, which 
%
%
plays a crucial role in capturing and processing information in hypercomplex networks, especially in complex tasks involving multidimensional signals with inherent rotation or spatial symmetries. In particular, the definition of neural network layers and learning operations in the hypercomplex domain allows capturing intricate inter-channel and intra-channel dependencies.

The dimensionality bias also potentially leads to more expressive and nuanced representations. Indeed, the multidimensional signals can be represented as a single entity in the hypercomplex domain. This is the case, for instance, of signals like color images or 3D audio signals \cite{EllTIP2007, ComminielloICASSP2019a}. 
%
%
%
%
%
\subsection{Algebraic Bias}
Hypercomplex numbers exhibit distinct algebraic structures that are different from those in real and complex numbers. Models operating in the hypercomplex domain could be biased towards leveraging these specific algebraic properties, such as the distributive, associative, and commutative properties of the underlying hypercomplex algebra. Such properties can be used as a bias to model and represent complex interactions between features that do not adhere to classic real-valued algebra rules. 
Depending on the specific hypercomplex algebra being considered (e.g., quaternions, octonions), algorithms can be designed to leverage the unique properties of these algebras, which represent the set of \textit{algebraic biases}.

For instance, in the quaternion domain, one algebraic property is the non-commutativity of the multiplication (i.e., $pq \neq qp$, being $p$ and $q$ quaternion numbers), which is the core of the Hamilton product. This property can be exploited to bias the learning process in neural networks through the operations in quaternion layers. The algebraic bias encourages the model to capture interactions and relationships that are influenced by the non-commutativity of quaternions. 
Figures~\ref{fig:qlayer} show a graphical representation of the quaternion layer, where a real-valued fully-connected layer does not show any particular inductive bias, while the sharing of the weight parameters in a quaternion layer is a strong bias even in such simple types of learning modules.

In octonion deep learning, the algebraic bias arises from the properties of octonion algebra, which include non-associativity of multiplication in addition to non-commutativity, i.e., $\left(pq\right)r \neq p\left(qr\right)$, with $p$, $q$ and $r$ as octonion numbers. 
This bias stems from the fact that different orders of operations can lead to different results, thus affecting the way neural networks using octonion-based representations learn and generalize.

As another example, in deep learning based on Clifford algebra, which generalizes real, complex and quaternion-valued algebras, the algebraic bias arises from the properties of Clifford multiplication and the ways these properties influence the model behavior. Clifford algebra is built up on a set of basis elements $\left\{\mathbf{e}_1, \mathbf{e}_2, \ldots, \mathbf{e}_n\right\}$, often represented as geometric vectors. These basis elements anticommute, i.e., $\mathbf{e}_i \land \mathbf{e}_j = - \mathbf{e}_j \land \mathbf{e}_i$, for $i \neq j$, where $\land$ denotes the Clifford multiplication that defines the algebraic structure. Neural network layers relying on Clifford multiplication are strongly affected by the anticommutation property, which biases the network to learn representations that align with the underlying structure of Clifford algebra, offering potential advantages in applications.

Algebraic biases present both challenges and opportunities. On one hand, they provide deep learning models with a unique way to capture intricate interactions and dependencies that may be difficult to represent using traditional number systems. On the other hand, they require careful consideration when designing network architectures and learning algorithms.
%
%
%
%
\subsection{Geometric Bias}
Hypercomplex numbers possess rich geometric properties, and these geometries can be leveraged to impose inductive biases on the learning process. The geometric structure of hypercomplex numbers can help capture meaningful information that might be difficult to express using traditional real numbers. Hypercomplex numbers are well suited to represent rotations and transformations in higher-dimensional spaces. When mapping data from a real or complex domain to a hypercomplex domain, deep learning algorithms can be biased towards maintaining the structural properties of the original data, ensuring meaningful transformations. Geometric biases affect the learning process by incorporating geometric principles into the model representations. 

Similarly to the algebraic properties, geometric biases can vary depending on the specific hypercomplex algebra adopted. Moreover, hypercomplex numbers allow for operations that are more closely aligned with geometric transformations. For example, the rotation of a vector can be then achieved by performing two Hamilton products among the polar quaternion, the vector, and the conjugate of the quaternion. 
This operation naturally combines both the geometric and algebraic aspects of hypercomplex numbers. 

Also, quaternions 
enable smooth interpolation between different orientations in a 3D space, making quaternion deep learning valuable for applications where spatial transformations are important.

Another hypercomplex algebra that shows interesting geometric biases is represented by dual-quaternion numbers, which can represent rigid body transformations, including both translation and rotation, in a very compact and elegant fashion. In fact, dual quaternions comprise two quaternion numbers, the latter multiplied by the dual unit, for a total of 8 degrees of freedom (DoF). By normalizing a dual quaternion to be unital, the DoF is reduced to 6, which can represent the three coordinates $x_r,y_r,z_r$ of the rotation in the first quaternion, and the $x_t,y_t,z_t$ coordinates for the translation in the dual part of the number \cite{GRASSUCCIPRL2023}. This enables dual-quaternion neural networks to deal with tasks involving very complex transformations in a high-dimensional space, like pose estimation, by simultaneously combining rotations and translations.

%
%
%
%
\subsection{Invariance and Equivariance Biases}
Invariance biases play a crucial role in deep learning by ensuring that model predictions or representations remain consistent despite certain transformations or changes in the input data. In the context of hypercomplex deep learning, invariance properties are particularly interesting as they leverage the geometric and algebraic properties of hypercomplex numbers. These properties can lead to novel ways of achieving and utilizing invariance in neural networks. 

Invariance to geometric transformations (e.g., rotation, translation, and scaling) can be explicitly built into a hypercomplex deep learning architecture. \textit{Geometric invariance} allows the model to capture significant features regardless of any orientation or position of the desired information (e.g., a specific object in an image). Biases can be induced by hypercomplex models also to exploit data topology 
by performing topology-preserving transformations 
while preserving the topological relationships between points.

\begin{figure}[t]
    \centering   
    \includegraphics[width=\textwidth]{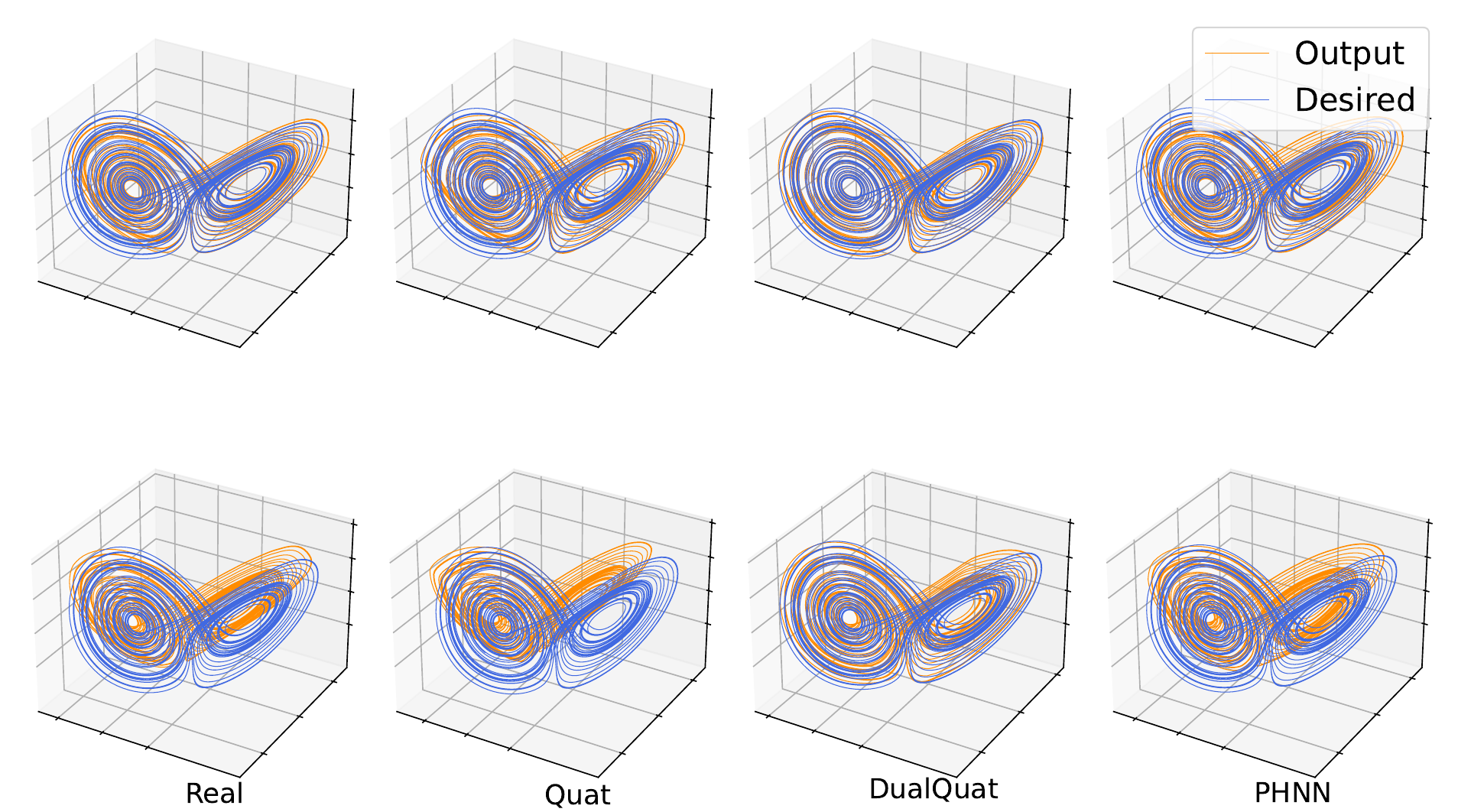}
    \caption{Models predicting the future position of points in a Lorenz system in the case of the original test set (first row) and the translated test set (second row). Real- and quaternion-valued neural networks fail to properly describe the system with translation, while the model equipped with a dual-quaternion underlying algebra faithfully predicts the translated system, demonstrating its translation-equivariant property
    \cite{VieiraMLSP2023}. In the last column, observe that PHNNs lose translation-equivariant benefits. Original image in \cite{VieiraMLSP2023}.}
    \label{fig:dualq_trans}
\end{figure}

Moreover, some hypercomplex number systems, like quaternions, can encode scale information. This property can be leveraged to build \textit{scale-invariant} representations. A hypercomplex network could learn to extract features that are invariant to changes in the scale of objects, which can be beneficial for tasks like object detection and image classification. Hypercomplex numbers are also well-suited for handling viewpoint changes. Models relying on hypercomplex representations can be designed to be \textit{perspective-invariant} to changes. This bias is crucial for tasks like 3D object recognition and scene understanding, where objects may be observed from various angles. For applications in image processing or computer vision, hypercomplex models could potentially be used to capture \textit{color invariance}. 
Also, \textit{representation invariance} can be introduced to improve model stability. For example, a hypercomplex network might learn to encode critical features of an object or scene in an invariant fashion, reducing the impact of noise or small perturbations.

Hypercomplex deep learning models can also benefit from equivariance biases. These properties are strictly related to geometric transformations in the hypercomplex domain. A typical example is given by the \textit{rotation equivariance} in quaternion neural networks. A 3D vector rotation is fully described by a rotation axis with a predefined angle. 
Scaling the angle while keeping the axis unchanged leads to
naturally disentangling a rotation-invariant element and a rotation-equivariant one \cite{QinTPAMI2022}. Neural networks, which possess this algebraic construction, therefore naturally inherit robustness to random and redundant rotations (such as those coming from different points of view of the object) and equivariance to informative rotations (those that are effectively applied to the object). 


Another example of hypercomplex equivariance bias is represented by the \textit{translation equivariance} in dual-quaternion neural networks. 
Owing to the explicit representation of the screw axis, dual quaternions are independent of the coordinate system, meaning that input data expressed in this way are translation-equivariant, different from quaternions whose rotations around the axes containing the origin are deformed by translations. This property naturally arises from the algebraic construction of dual quaternions, and it is illuminated in Fig.~\ref{fig:dualq_trans}, on a toy problem \cite{VieiraMLSP2023}. 

%
%
%
%
\subsection{Regularization Bias}
Similarly to the real-valued case, hypercomplex deep learning may induce some biases based on non-relational assumptions, i.e., those constraints, or priors, that guide the learning process and representation of signals without explicitly relying on relational information or pairwise interactions between input data points. Hypercomplex deep learning models can exhibit non-intuitive behaviors, e.g., due to non-commutativity, which can lead to instability during learning. Deep learning models can incorporate regularization techniques tailored to the hypercomplex domain to ensure stability and convergence, control the model complexity, prevent overfitting phenomena, and learn representations that are both informative and robust. The set of regularization approaches that can be used in hypercomplex deep learning can be referred to as \textit{regularization bias}.

A typical non-relational regularization bias in deep learning is represented by penalty terms in loss functions that encourage certain properties or constraints on the model weights or activations. When dealing with such regularization terms in the hypercomplex domains, the properties of the specific algebras must be taken into account. 
Another non-relational regularization bias is represented by data augmentation. Augmentation techniques in the hypercomplex domain rely on geometric transformations, like translation, rotation, and flipping, which enable the model to modify the non-relational aspects of input signals without altering their inherent meaning, thus enhancing the model ability to generalize. Moreover, data augmentation may benefit from invariance properties to lead to more robust hypercomplex models.

Differently from real-valued deep learning regularization assumptions, which are basically non-relational, the regularization bias for the hypercomplex domain can also include relational properties that mainly rely on some specific operations like the Hamilton product. Indeed, the Hamilton product can serve as a mechanism to regulate model complexity, thus ensuring that the network focuses on learning relevant features and interactions without becoming overly complex. 
It is worth noting that while the Hamilton product introduces regularization-like effects in hypercomplex deep learning, it does not replace traditional regularization techniques but rather complements them. In practice, non-relational regularization approaches can be used in combination with the Hamilton product to achieve the desired balance between model complexity, generalization, and robustness. However, it must be considered that any non-relational regularization technique applied to hypercomplex models will not bring the same performance improvement gains as those applied to real-valued models, precisely because hypercomplex models are less affected by overfitting owing to the regularizing effect of Hamilton product \cite{GrassucciTNNLS2022}.
%
%
%
%
%
\section{Inductive Bias Beyond Hypercomplex Algebras}
\label{sec:phnninductive}
Hypercomplex deep learning models strongly benefit from domain-specific inductive biases. However, such specificity may often limit the range of possible applications for which bias assumptions hold. For instance, not all the hypercomplex algebras can take advantage of the non-commutativity property of the Hamilton product, typical of the quaternion domain. Moreover, not all the dimensionality orders $n$ are endowed with known algebras. In order to overcome these limits, it is possible to go beyond the Cayley-Dickson construction of well-defined hypercomplex algebras by parameterizing hypercomplex layers \cite{Zhang2021PHM, GrassucciTNNLS2022}. Parameterized hypercomplex neural networks (PHNNs) are able to force some assumptions beyond a specific algebra, thus extending the diversity of the multidimensional signals that can be processed.
%
%
%
%
\subsection{Parameterized Hypercomplex Deep Learning}
Similar to Section~\ref{sec:hypdom}, we can define a set of learning operations for PHNNs. The core of parameterized hypercomplex layers relies on the Hamilton product of quaternions that can be decomposed into a sum of four Kronecker products between the matrices that express the algebra rules and the actual layer weights. This allows for the preservation of the property of weight sharing that is at the core of quaternion neural networks. To this end, a weight matrix $\mathbf{W}$ can be defined as:
\begin{equation}
\label{eq:PHweight}
    \mathbf{W} = \sum_{i=1}^n \mathbf{A}_i \otimes \mathbf{F}_i,
\end{equation}

\noindent where $\otimes$ represents the Kronecker product, $\mathbf{A}_i \in \bR^{n \times n}$ with $i=1, ..., n$ are the learnable matrices that grasp the algebra rules from data and organize weights/filters for fully connected or convolutional layers, $\mathbf{F}_i \in \bR^{\frac{s}{n} \times \frac{d}{n}}$ for FC layers and $\mathbf{F}_i \in \bR^{\frac{s}{n} \times \frac{d}{n} \times k \times k}$ for convolutions, are the batches of weights/filters that are arranged by $\mathbf{A}_i$ to compose the final weight matrix, and $n$ is any (i.e., not limited to known domains) integer scalar hyperparameter that defines the domain dimensionality. 

PHNNs can define the most common learning operations, starting from the parameterized hypercomplex multiplication (PHM) layer:
\begin{equation}
\label{eq:phm}
    \y = \text{PHM}(\x) = \mathbf{W}\otimes\x + \mathbf{b},
\end{equation}

\noindent and the parameterized hypercomplex convolutional (PHC) layer with $\mathbf{W} \in \bR^{s \times d \times k \times k}$ for 2D convolutions:
\begin{equation}
\label{eq:phc}
    \y = \text{PHC}(\x) = \mathbf{W}*\x + \mathbf{b}.
\end{equation}

Notably, PHC layers allow processing RGB images in their natural domain by easily setting $n=3$ without padding a zero channel as is usually performed in quaternion and tessarine neural networks, thus enhancing model performance. 

Also, parameterized hypercomplex self-attention (PHAtt) layers can be defined following \cite{Zhang2021PHM}:
\begin{equation}
\begin{aligned}
    \mathbf{Q}, \mathbf{K}, \mathbf{V} &= \phi(\text{PHM}(\x)), \\ 
    \text{PHAtt} &= \text{softmax} \left( \frac{\mathbf{Q}\mathbf{K}^\top}{\sqrt{d_k}} \right) \mathbf{V}, \\
    \mathbf{y} &= \text{PHAtt} \cdot \x .
    \label{eq:phatt}
\end{aligned}
\end{equation}

\noindent where $d_k$ is the key dimension; multi-head attention can also be implemented by concatenating multiple heads before processing them with further PHM layers \cite{Zhang2021PHM}.

In a very similar way as in generic hypercomplex numbers, parameterized hypercomplex aggregation operator for graphs can be defined as
\begin{equation}
    \mathbf{w}_v^{l+1} = \phi \left( \Sigma_{u \in \mathcal{N}_v \bigcup\{v\}} a_{v,u} \mathbf{W}^l \otimes \mathbf{w}_u^l \right), \qquad \forall v \in \mathcal{V},
\end{equation}

\noindent where $\mathbf{W}$ is built according to \eqref{eq:PHweight}. Moreover, specific input factorizations have been proposed to encode graph features into hypercomplex numbers, together with message passing operators \cite{LeICANN2021}.
%
%
%
%
\subsection{Inductive Biases in Parameterized Hypercomplex Models}
The formulation of parameterized hypercomplex operations significantly affects the biases defined in Section~\ref{sec:hyperinductive}.

\begin{figure}[t]
    \centering
    \includegraphics[width=0.6\linewidth]{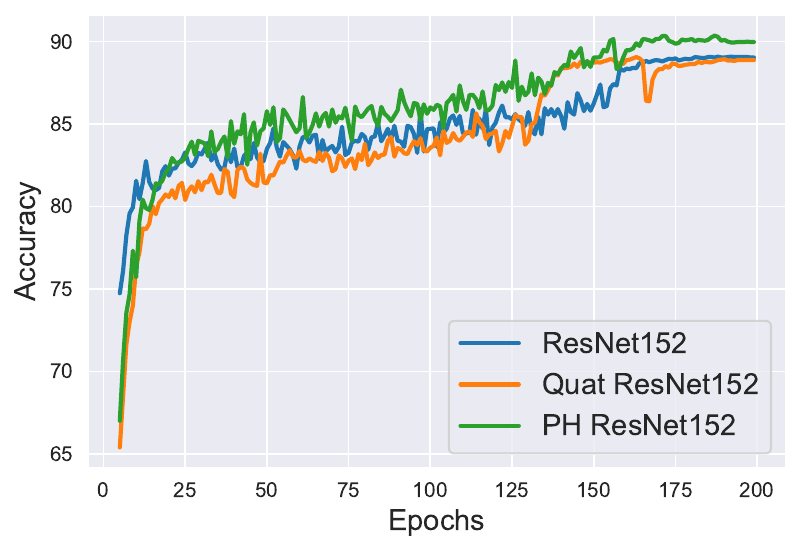}
    \caption{PH ResNet outperforms competitors in CIFAR10 classification while comprising $1/3$ parameters of the conventional ReNet.}
    \label{fig:phnn_class}
\end{figure}

\begin{figure}[t]
    \centering
    \includegraphics[width=\linewidth]{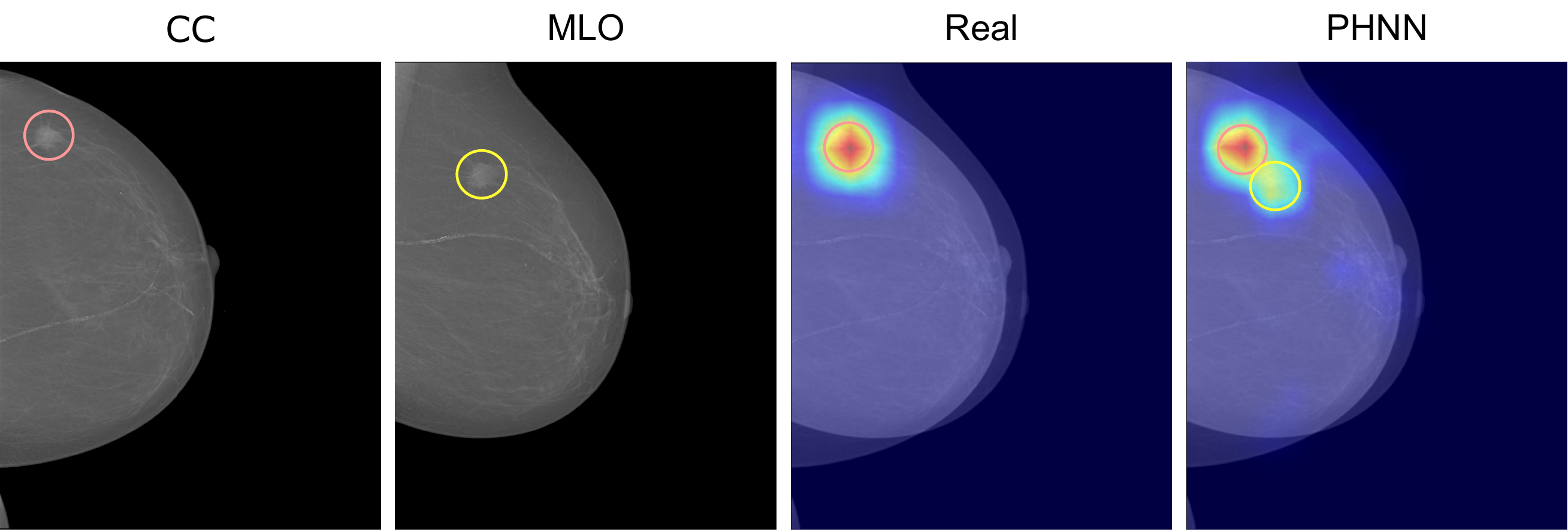}
    \caption{When processing two-view mammogram exams, the real-valued model focuses on a single view only, while the PHNN crucially captures both-view information, leading to a more robust detection. Original image in \cite{LopezARXIV2023}.}
    \label{fig:gradcam_breast}
\end{figure}

\textbf{Dimensionality bias.} 
This bias is extremely enhanced by parameterized hypercomplex operations. Indeed, the real potential of PHNNs lies in the extension of the dimensionality biases to domains for which algebra rules are unknown. While it is possible to define the domain in which the model operates by easily setting the hyperparameter $n$, this parameter clearly refers to the dimensionality of an input signal. Thus, PHNNs can operate with any $n$-dimensional signal. In fact, in PHNNs both algebra matrices, $\mathbf{A}_i$, and adaptive weights, $\mathbf{F}_i$, are learnable, thus bringing a breakthrough in the field of hypercomplex deep learning. Indeed, PHNNs generalize pre-existing hypercomplex networks by extending their advantages to any $n$D domain, regardless of whether the algebra rules are known. Therefore, e.g., by setting $n=3$ the model can process RGB images in their natural domain, thus leveraging the weight-sharing property and learning the hypercomplex rule for multiplication directly from data. This offers enhanced results compared to conventional real-valued models that can be seen in the classification example in Fig.~\ref{fig:phnn_class}, where the PH ResNet152 outperforms competitors even though it is defined with $1/3$ of the parameters. Similarly, PHNNs can equally exploit complementary information in multiview data due to the dimensionality bias, while real-valued counterparts focus on single views only losing crucial information as in Fig.~\ref{fig:gradcam_breast}.

\textbf{Algebraic bias.} 
The flexibility of PHNNs leads to redefining the algebraic bias, as the model loses the domain-specific properties while exploiting the Hamilton-like structure of the PHM in \eqref{eq:phm}.
The processing of multidimensional signals can now benefit from the weight-sharing ability of the Hamilton multiplication regardless of the specific dimensionality considered, thus always preserving the local relations among the input dimensions. By exploiting the Hamilton-like multiplication, we force the model to extend its learning ability to any dimensionality and to grasp algebra rules directly from input signals, organizing weights regardless of whether the algebra is preset. This enables the processing of a wide variety of multidimensional signals in the hypercomplex domain. Indeed, the PHM weight sharing brings improved results in several tasks such as image classification or sound detection \cite{GrassucciTNNLS2022}. 

\textbf{Regularization bias.} 
The Hamilton-like definition of the PHM also preserves the relational aspects of the regularization bias that relies on parameter reduction and sharing. The effect of such regularization is clearly evident when dealing with large-scale deep learning models, where huge numbers of parameters may introduce significant overfitting while showing impressive performance. In those cases, a re-definition of the model in the hypercomplex domain can lead to achieve similar impressive performance, while significantly reducing the number of parameters \cite{GrassucciTNNLS2022}.

\textbf{Geometric-related biases.} 
It should be noted that the loss of domain-specific geometric benefits, due to the flexible definition of the underlying algebra rules, significantly affects the invariance and equivariance biases. In particular, PHNNs do not preserve the rotation and translation equivariant properties in the 3D space that remain pure biases respectively related to quaternion and dual quaternion models, as Figure~\ref{fig:dualq_trans} shows.

\textbf{Recovering domain-specific inductive biases.} 
PHNNs generalize hypercomplex neural networks; indeed, it is possible to collapse the model into any hypercomplex algebra and recover all the domain-specific inductive biases. Let us consider the case of the quaternion domain as an example. By fixing the four algebra matrices $\mathbf{A}_i, \; i=1, ..., 4$ (one for each quaternion coefficient) to represent the Hamilton product, the resulting layer also preserves the geometry-related properties of the quaternion domain, such as rotation equivariance, since no alteration is applied to the algebra of the layer. 
This formulation penalizes the flexibility of PHNNs while recovering the domain-specific biases.
%
%
%
%
%
%
%

\section{Application Examples for Multidimensional and Multimodal Signals}
\label{sec:applications}
The choice of inductive biases in a certain hypercomplex deep learning model should be guided by the specific characteristics and goals of the application problem at hand.
Different applications may require different biases to effectively leverage the benefits of the hypercomplex domain.

Moreover, sometimes it is possible to choose different ways to process signals in the hypercomplex domains according to the kind of inductive bias that better generalizes a hypercomplex deep learning model in a specific application. The most popular approach is to leverage the dimensionality bias by embedding multidimensional signals into the hypercomplex axes. For instance, it is possible to encapsulate the R-G-B channels of a color image into the imaginary axes of a pure quaternion. The hypercomplex model that processes this kind of input will benefit from the inter and intra-channel correlations. A second approach, which is recently becoming more frequent in emerging applications, is more suited for leveraging geometric-related biases. The idea is to represent a complex input signal in a low-dimensional latent space, where it is easy to identify a number of significant characteristics that represent the input signal by their shared semantics \cite{GuizzoTASL2023, WangKBS2023}.

\textbf{Image Analysis and Synthesis.} One of the most popular categories of applications where it is possible to exploit the abilities of several hypercomplex algebras is image analysis and synthesis.
Besides exploiting the classic deep learning assumptions (e.g., convolutional networks inductive biases), hypercomplex deep learning applied to images easily allows to leverage the dimensionality bias, as color image channels exhibit a high intra-channel correlation. Since the dimensionality of the images is often 3D, the hypercomplex algebra that is most adopted with this kind of signal is the quaternion one \cite{Yang2023QTN}. Quaternion neural networks are ideally suited for images as they can bias the learning with both algebraic and geometric-related assumptions, as well as several regularization biases. The same properties also hold in image synthesis, thus hypercomplex deep generative models can be easily built to deal with images \cite{GrassucciQGAN2021, GrassucciICASSP2021}.
Similarly, going beyond pre-defined algebras, images are well processed by PHNNs as they can handle images in their natural domain thus leading to improved results as in Fig.~\ref{fig:phnn_class}.

\textbf{3D Audio.} Hypercomplex deep learning has been widely applied to multichannel audio processing \cite{ComminielloICASSP2019a}. 
In detail, quaternion neural networks are particularly suited to deal with first-order Ambisonics 3D audio signals \cite{BrignoneTCAS2022}, for which all the quaternion-domain inductive biases can be easily exploited. In 3D audio scenarios, geometric biases like quaternion rotations may prevent ambiguities in estimating the directions of arrival of sound signals and can enable a fast virtual miking \cite{ComminielloICASSP2019a}. 
Moreover, extensions to high-order or arrays of Ambisonics signals can be performed by exploiting the properties of dual-quaternion and parameterized hypercomplex algebras \cite{GRASSUCCIPRL2023, GrassucciTNNLS2022}. In particular, dual-quaternion neural networks allow to exploit rotation and translation equivariance of sound sources in 3D acoustic environments. 

\textbf{Speech emotion recognition.} Hypercomplex deep learning can be applied also to monodimensional signals. In those cases, some transformations of the input is required to embed the input signal into a hypercomplex representation. One example is speech emotion recognition, where speech can be represented in terms of valence, arousal, dominance, and overall emotion \cite{GuizzoTASL2023, Muppidi2021ICASSP}. These characteristics may represent the multidimensionality of a latent space, where hypercomplex neural networks can be applied to perform classification, reconstruction or even generation of speech emotions. In this scenario, geometric-based biases are significantly exploited. Rotations and translations in the hypercomplex latent space denote a change in emotions along the speech emotion characteristic axes. Invariance and equivariance biases can help to preserve specific aspects of a speech emotion.

\textbf{Time series analysis and forecasting.} Hypercomplex deep learning has been widely adopted in the literature to deal with time series. Several models have been built to leverage the dimensionality bias, thus embedding multichannel/multivariate time series in the hypercomplex components to benefit from the input correlations (e.g., \cite{BallanteNAT2023}). 
Another way to generalize hypercomplex models in time series analysis is to work in hypercomplex latent spaces where geometric transformations bias the models. For instance, in \cite{ChenKDD2022}, a learning-to-rotate attention mechanism was proposed to depict intricate periodical patterns, thus effectively learning multiple and variable periods and phase information.

\textbf{Natural language processing.} Hypercomplex deep learning models are also well suited for NLP tasks, such as pairwise text classification, question answering, sentiment analysis, text style transfer, and neural machine translation \cite{Tay2019QTRansformer, Zhang2021PHM}. In these tasks, each word embedding is treated as a concatenation of hypercomplex components. Dimensionality bias in this case helps grasping the semantic correlation in the input sequences, while the algebraic bias is exploited in the definition of the hypercomplex deep learning model, such as the quaternion transformer \cite{Tay2019QTRansformer}.

\textbf{Multimodal and multi-view signal processing.} Hypercomplex neural networks can offer insights into complex cross-modal relationships that may be challenging to capture using other methods. Multimodal signals can benefit from geometric-related biases via transformations to understand the patterns representing the shared semantics of an input \cite{WangKBS2023}. Moreover, the algebraic biases force a hypercomplex model to capture the asymmetric relevance. This also happens for multi-view signal analysis, like breast cancer screening, where a hypercomplex model is able to leverage the intrinsic asymmetries in breast screening views to better detect possible cancers, thus mimicking clinicians reading process \cite{LopezARXIV2023, LopezPRLetters2024}.

\textbf{Point data modeling.} Several kinds of data, such as skeletons and point clouds, can be represented by points in 3D space.
These data exhibit a natural topological structure and can be easily embedded in quaternions or other hypercomplex numbers just by considering the point coordinates and relying on the dimensionality and algebraic biases to perform inference. However, they can also be represented by hypercomplex numbers whose components refer to a rotation or to a particular geometric transformation of each point in the 3D space \cite{QinTPAMI2022}. Several neural networks, including transformers, graph neural networks or message-passing models, can be implemented in the hypercomplex domains to take advantage of the possible biases that can be defined upon these data. Moreover, invariance and equivariance properties can be also leveraged by quaternion and dual-quaternion networks for complex tasks like action recognition and pose estimation in a higher-dimensional space\cite{QinTPAMI2022, VieiraMLSP2023}.

%
%
%
%
%
\section{Conclusion and Future Perspectives}
\label{sec:conclusion}
In this paper, we presented a novel perspective on hypercomplex deep learning under the lens of the inductive biases brought by hypercomplex algebras to the learning process of neural models defined in hypercomplex domains. The work aims to demystify hypercomplex deep learning models and make them suitable for a wide variety of multidimensional signal processing applications. We believe this work may pave the way for new discoveries new scientific discoveries in deep learning. In particular, future perspectives may definitely include the investigation of biases for the development of novel hypercomplex deep learning models that exploit at best the favorable algebraic rules. In addition, the biases analyzed in this paper may be leveraged for emerging applications and for the development of novel generalizable frameworks that wisely combine multiple biases (e.g., algebraic and geometric) to produce a comprehensive data representation that better describes the intrinsic physical process that generated it.
%
%

\section*{Acknowledgement}
This work was supported by the European Union under the Italian National Recovery and Resilience Plan (NRRP) of NextGenerationEU, partnership on ``Future Artificial Intelligence Research" (PE0000013 - FAIR - Spoke 5: High Quality AI) and on ``Digital Driven Diagnostics, prognostics and therapeutics for sustainable Health care" (PNC 0000001 - D3 4 Health - Spoke 1: Clinical use cases and new models of care supported by AI/E-Health based solutions). This work was also supported by the Italian Ministry of University and Research (MUR) within the PRIN 2022 Program for the project ``EXEGETE: Explainable Generative Deep Learning Methods for Medical Signal and Image Processing", under grant number 2022ENK9LS. This work was also partially supported by the ``Progetti di Avvio alla Ricerca", under grant AR2221813925AB35.
%
%
%
%
\bibliographystyle{IEEEtran}
\bibliography{SPMbib}
%
%
%
%
%
\section*{}
\textbf{Danilo Comminiello} received the Ph.D. degree in Information and Communication Engineering in 2012 from Sapienza University of Rome, Italy. He is currently an Associate Professor with the Department of Information Engineering, Electronics, and Telecommunications (DIET) of Sapienza University of Rome, Italy. He is an elected Member, and currently Vice-Chair, of the IEEE Machine Learning for Signal Processing Technical Committee.

\textbf{Eleonora Grassucci} received the Ph.D. degree in Information and Communication Technologies in 2023 from Sapienza University of Rome, Italy. She is an Assistant Professor at the Department of Information Engineering, Electronics, and Telecommunications of the Sapienza University of Rome. She was awarded the Best Track Manuscript Recognition by the IEEE Circuits and Systems Society in 2022 and the Doctoral Dissertation Award by the International Neural Networks Society in 2023.

\textbf{Danilo P. Mandic} received the Ph.D. degree in nonlinear adaptive signal processing from Imperial College London, London, U.K., in 1999. He is currently a professor in signal processing at Imperial College London. He is a guest professor with the
Katholieke Universiteit Leuven, Leuven, Belgium, Tokyo University of Agriculture and Technology, Tokyo, Japan, and a frontier researcher with RIKEN, Japan. He was the recipient of the Denis Gabor Award for Outstanding Achievements in Neural Engineering, given by the International Neural Networks Society, and the winner of the 2018 Best Paper Award in IEEE Signal Processing Magazine.

\textbf{Aurelio Uncini} received the Ph.D. degree in Electrical Engineering in 1994 from the University of Bologna, Italy. Currently, he is a Full Professor at the Department of Information Engineering, Electronics, and Telecommunications, where he is teaching Neural Networks, Adaptive Algorithms for Signal Processing and Digital Audio Processing, and where he is the Head of the ``Intelligent Signal Processing and Multimedia'' (ISPAMM) group.

\end{document}